\documentclass[journal,twoside,web]{ieeecolor}
\usepackage{generic}
\usepackage{amsmath,amssymb,amsfonts}
\usepackage{algorithmic}
\usepackage{graphicx}
\usepackage{algorithm,algorithmic}
\usepackage{hyperref}
\usepackage{textcomp}
\usepackage[capitalise]{cleveref}
\usepackage{placeins}
\usepackage{subfig}
\usepackage{stfloats}
\usepackage{booktabs}
\usepackage{adjustbox}
\usepackage{float}
\usepackage{multirow}

\usepackage{cite}

\begin{document}

\title{
    \textbf{
        \parbox{0.8\textwidth}{\centering 
        \fontsize{18pt}{20pt}\selectfont 
        uPVC-Net: A Universal Premature Ventricular Contraction Detection Deep Learning Algorithm}
    }
}

 
\author{Hagai Hamami, Yosef Solewicz, Daniel Zur, Yonatan Kleerekoper and Joachim A. Behar, \IEEEmembership{Senior Member, IEEE} \thanks{Manuscript submitted on \today. Research was supported by Israel PBC-VATAT. (Corresponding author: Joachim A. Behar, jbehar@technion.ac.il). We acknowledge the assistance of ChatGPT, an AI-based language model developed by OpenAI, in editing the manuscript.} \thanks{H. Hamami, Y. Solewicz and J. A. Behar are with the Faculty of Biomedical Engineering, Technion-IIT, Haifa, Israel. D. Zur and Y. Kleerekoper are with the The Ruth and Bruce Rappaport Faculty of Medicine, Technion-IIT, Haifa, Israel.}
}

\maketitle

\begin{abstract}
Introduction: Premature Ventricular Contractions (PVCs) are common cardiac arrhythmias originating from the ventricles. Accurate detection remains challenging due to variability in electrocardiogram (ECG) waveforms caused by differences in lead placement, recording conditions, and population demographics. Methods: We developed uPVC-Net, a universal deep learning model to detect PVCs from any single-lead ECG recordings. The model is developed on four independent ECG datasets comprising a total of 8.3 million beats collected from Holter monitors and a modern wearable ECG patch. uPVC-Net employs a custom architecture and a multi-source, multi-lead training strategy. For each experiment, one dataset is held out to evaluate out-of-distribution (OOD) generalization. Results: uPVC-Net achieved an AUC between 97.8\% and 99.1\% on the held-out datasets. Notably, performance on wearable single-lead ECG data reached an AUC of 99.1\%. Conclusion: uPVC-Net exhibits strong generalization across diverse lead configurations and populations, highlighting its potential for robust, real-world clinical deployment.

\end{abstract}



\section{Introduction}
\label{sec:introduction}

\IEEEPARstart{P}{remature} ventricular contraction (PVC) is a common type of abnormal heartbeat that originates from an ectopic focus within the ventricles. The frequency and morphology of PVC were shown to be associated with a reduced left ventricular ejection fraction (LVEF), heart failure\cite{dukes_pvc_HF_death_2015}, cardiomyopathy~\cite{baman2010relationship}, stroke\cite{agarwal_pvc_stroke_2011,agarwal_pvc_stroke_2016} and increased mortality\cite{yu_multiform_pvc_2014}.

There is compelling evidence demonstrating not only the independent association between high PVC burden and reduced LVEF~\cite{baman2010relationship}, but also the reversibility and LVEF normalization with PVC suppression through catheter ablation~\cite{bogun2007radiofrequency, latchamsetty2015multicenter, penela2017clinical}. Therefore, early detection of patients with a high PVC burden and prompt cardiac evaluation may facilitate timely, targeted interventions and potentially prevent the progression to PVC-induced cardiomyopathy~\cite{baman2010relationship}. Embedding such algorithms into wearable devices could enable continuous assessment of PVC burden and facilitate long-term monitoring of treatment efficacy, such as PVC suppression, as well as early detection of complications, including the emergence of new ectopic foci.

To further investigate the association between PVC and clinical outcomes in large-scale ECG datasets, as a first step towards any integration into clinical practice, highly robust and reliable PVC detection algorithms are essential.

A significant body of research has focused on improving PVC beat detection using machine learning. The evolution of PVC detection methods has progressed from traditional rule-based and feature-engineered approaches\cite{12-cai2019rule-based} to machine learning-based classification\cite{9-hajeb-mohammadalipour2018automated,8-oliveira2019geometrical,7-de2022classification} and, more recently, deep learning based algorithms\cite{10-sarshar2022premature,17-cai2022robust,18-ullah2022automatic,19-kraft2023end-to-end,11-guo2024pvcsnet}. Among these studies, only a few works \cite{24-llamedo2011heartbeat,25-li2019ventricular,26-cai2024hybrid} reported out-of-distribution generalization performance (OOD-GP) in an external dataset(s). Out of which, only the work of Li et al. \cite{25-li2019ventricular} was identified as reporting OOD-GP on an open access dataset (MIT-BIH), and doing so without excluding any type of beat from the analysis.

Achieving strong generalization in physiological time-series analysis remains a substantial challenge due to the complexity and variability of human biological signals. While deep learning models have demonstrated impressive results on specific datasets of physiological measurements~\cite{phan_seqsleepnet_2019,levy_deep_2023,ribeiro_automatic_2020}, their performance often drops significantly when evaluated on external data~\cite{kotzen_sleepppg-net_2023,levy_deep_2023,ballas_towards_2024}. This limitation goes beyond a technical concern since it hinders the translation of these methods into real clinical applications. Consequently, there is a critical need for developing models that not only excel on a held out test set of the development dataset but also remain robust across diverse population samples and sensor technology~\cite{behar_generalization_2025}.

To address this challenge, our recent research has demonstrated the effectiveness of using multiple datasets to improve OOD-GP\cite{attia_sleepppg_2024,men_deep_2025,abramovich_gonet_2025,eran_multisource_2025}. In addition, in ECG analysis, we have shown that training deep learning models on multiple lead positions from a single dataset improves their ability to generalize to alternative lead placements\cite{ben-moshe_rawecgnet_2024}. Building on these previous findings, the present work aims to develop a universal PVC beat detector capable of accurately detecting PVC beats from single-lead ECG recordings across various lead positions, diverse population samples, and ECG sensing technology, including wearable. By incorporating a multi-source domain multi-lead training strategy along with a custom deep learning architecture and preprocessing steps, we seek to create a robust and adaptable model for clinical use. 



This research begins by introducing the architecture and strategy used to train uPVC-Net, a deep learning model designed for universal PVC beat detection. The proposed approach leverages a multi-source-domain multi-lead training strategy to improve OOD-GP in ECG analysis. A series of experiments are then conducted to evaluate the model’s OOD-GP across distribution shifts, using four independent open-access datasets totaling 8.3 million beats, originating from different geographical regions, demographics, lead positions and hardware types, ensuring the model’s robustness in diverse settings. Finally, a detailed quantitative error analysis is performed to identify major sources of false positives (FP) and false negatives (FN), improving interpretability and clinical reliability. The key contributions of this study are summarized as follows:
\begin{itemize} 
\item uPVC-Net, a novel deep learning model for PVC beat detection from raw ECG recordings; 
\item Evaluation of OOD-GP on multiple independent datasets;
\item A quantitative error analysis, identifying the primary sources of FP and FN.
\end{itemize}


\begin{figure*}[htbp]
    \centering
    \includegraphics[width=1\linewidth]{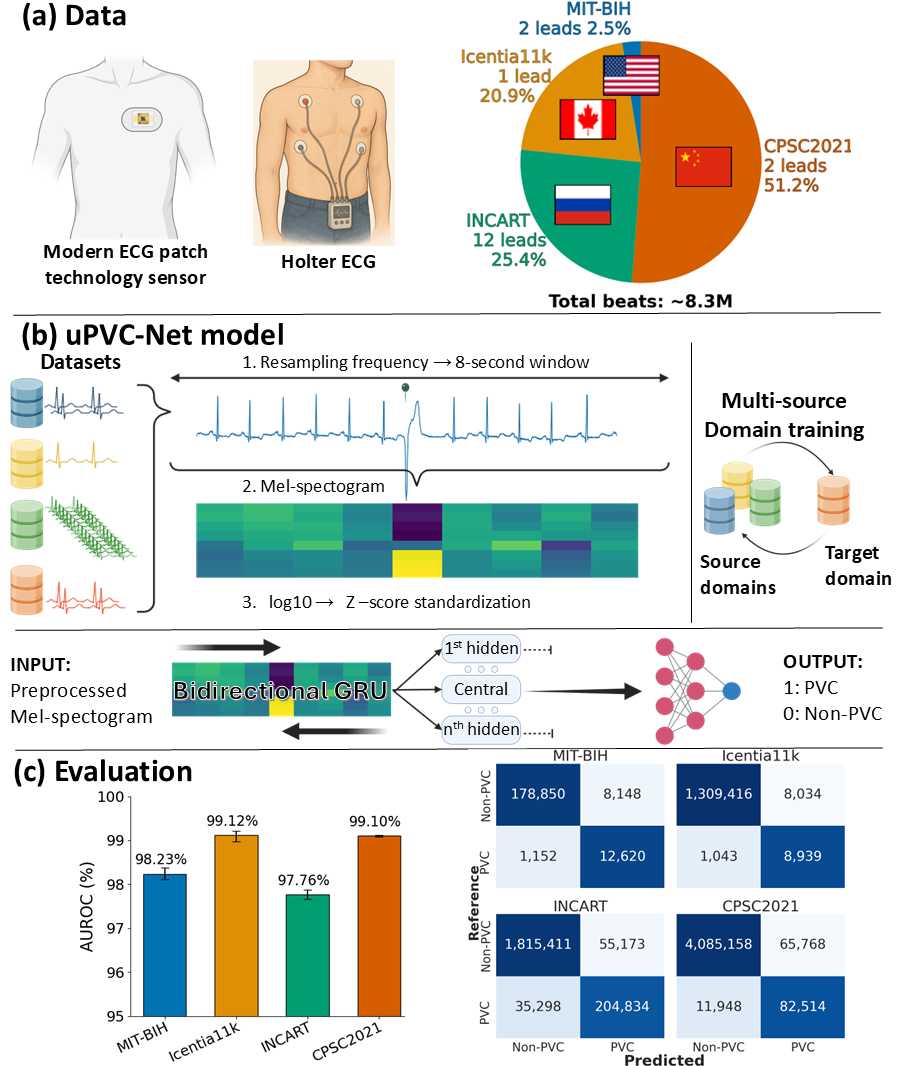}
    \caption{Research overview. (a) ECG recordings from both traditional Holter monitors and modern patch-based ECG sensors are included in the study. In total, four independent datasets from diverse geographical regions are utilized, comprising 8.3 million beats. (b) To train uPVC-Net, an 8-second ECG strip centered around a reference annotation serves as input to a deep neural network after being subjected to a series of transformations. A multi-source, multi-lead training strategy is employed to ensure robust model performance. (c) Out-of-distribution generalization performance of uPVC-Net is evaluated across all included datasets and compared against the benchmark.
}
    \label{fig:mainfigure}
\end{figure*}


\section{Materials}

\begin{table*}[htbp]
\centering
\noindent\textbf{}
\vspace{0.5em}
\resizebox{\textwidth}{!}{%
    \begin{tabular}{@{}lccccc@{}}
    \toprule
\textbf{} & \textbf{MIT-BIH} & \textbf{Icentia11k} & \textbf{INCART} & \textbf{CPSC2021} & \textbf{Total} \\ \midrule
Non-PVC, n (\% in dataset) & 186,998 (90.44\%) & 1,317,450 (76.09\%) & 1,870,584 (88.62\%) & 4,150,926 (97.66\%) & 7,525,958 (90.68\%)\\PVC, n (\% in dataset) & 13,772 (6.66\%) & 9,982 (0.58\%) & 240,132 (11.38\%) & 94,462 (2.22\%) & 358,348 (4.32\%)\\Unlabeled, n (\% in dataset) & 6,006 (2.9\%) & 403,994 (23.33\%) & 108 (0.01\%) & 4,834 (0.11\%) & 414,942 (5.0\%)\\Total beats, n (\% in all data) & 206,776 (2.5\%) & 1,731,426 (20.9\%) & 2,110,824 (25.4\%) & 4,250,222 (51.2\%) & 8,299,248 (100.0\%)\\Distinct heartbeats, n (\% in all data) & 103,388 (2.5\%) & 1,731,426 (41.9\%) & 175,902 (4.3\%) & 2,125,111 (51.4\%) & 4,135,827 (100.0\%)\\Geography & USA & Canada & Russia & China & -\\Patients, n & 43 & 330 & 32 & 105 & 510\\Records, n & 44 & 330 & 75 & 1425 & 1874\\Channels, n & 2 & 1 & 12 & 2 & -\\Sampling frequency, Hz & 360 & 250 & 257 & 200 & -\\Recording device & Del Mar Avionics model 445 & CardioSTAT by Icentia & 12-lead devices & 12-lead or 3-lead devices & -
\\ \bottomrule
\end{tabular}%
}
\captionof{table}{Holter ECG Datasets used for the experiments. Four recordings (102, 104, 107, and 217) are excluded from the MIT-BIH Arrhythmia due to paced beats, as well as 11 recordings from CPSC2021 due to the unavailability of the raw data.}
\label{tab:datasets}
\end{table*}

A total of four standard long-term monitoring ECG datasets were used for the experiments (Figure \ref{fig:mainfigure}.a): MIT-BIH Arrhythmia Database (MIT-BIH; USA)\cite{physiobank,mit_bih}, Icentia11k Database (Canada)\cite{physiobank,icentia11k_ref2,icentia11k_ref1}, St. Petersburg Institute of Cardiological Technics 12-lead Arrhythmia Database (INCART; Russia)\cite{physiobank} and The 4th China Physiological Signal Challenge 2021 Database (CPSC2021; China)\cite{physiobank,CPSC2021}. The overall dataset totals 8.3 million beats. Table \ref{tab:datasets} summarizes the number of PVC and non-PVC beats available in each of the datasets. Heartbeats were divided into PVC and non-PVC (two classes). Non-PVC beats included non-ectopic, supraventricular ectopic and fusion beats. Unlabeled beats were excluded from all experiments.

\textbf{MIT-BIH} The MIT-BIH Arrhythmia dataset\cite{mit_bih,physiobank} consists of 48 two-channel ambulatory ECG recordings, each lasting 30 minutes, obtained from 47 subjects. The dataset was collected between 1975 and 1979 at Beth Israel Hospital in Boston. Each beat was independently annotated by at least two experienced cardiologists and all disagreements were resolved to produce reference annotations. The ECG signals were digitized at 360 Hz with 11-bit resolution over a 10 mV range. Four recordings (102, 104, 107, and 217) contained paced beats and were excluded from the analysis, as recommended by ANSI-AAMI (1998)\cite{23-ec571998testing}. ECG signals from the first channel were mainly from the modified limb lead II (MLII) while signals from the second channel were mainly from lead V1. However, in five records different lead configurations were used. This dataset serves as a standard reference for the detection of arrhythmias and is widely used in the development and validation of machine learning models for PVC detection\cite{24-llamedo2011heartbeat,25-li2019ventricular,26-cai2024hybrid}.

\textbf{CPSC2021} The CPSC2021 dataset\cite{physiobank,CPSC2021} consists of 1436 variable-length ECG recordings on both lead I and lead II, extracted from either a 3-lead or a 12-lead ECG monitoring device, and sampled at 200 Hz. 11 recordings were missing from the PhysioNet repository. All 105 patients were also labeled as either atrial fibrillation patients (n=49) or non-atrial fibrillation patients (n=56). 

\textbf{INCART} The St. Petersburg INCART 12-lead Arrhythmia dataset\cite{physiobank} comprises 75 annotated ECG recordings, each lasting 30 minutes and sampled at 257 Hz. Each recording includes the standard 12-lead ECG configuration. The signals were digitized with 16-bit resolution. Beat annotations in the INCART database were initially generated using an automated detection algorithm, which placed annotations at the center of the QRS complex using data from all 12 leads. The annotations then underwent manual correction. However, while the beat classification was corrected, the annotation locations were not manually adjusted, which may result in occasional misalignment.

\textbf{Incentia11k} The Icentia11k dataset\cite{physiobank,icentia11k_ref2,icentia11k_ref1} comprises continuous raw ECG signals from 11,000 patients, totaling approximately 2 billion annotated beats. The raw ECG signals were subsequently segmented into 70-minute recordings (up to 50 segments per patient). For the purpose of our research and given the large size of this dataset, we randomly sampled one 70-minute recording per 330 individual patients. Data were collected using the CardioSTAT device, a single-lead heart monitor, recording at 250 Hz with 16-bit resolution, with a duration of up to two weeks. Beat annotations were performed by a team of 20 certified technologists, who labeled each beat as normal, premature atrial contraction, or PVC, and identified rhythms such as normal sinus rhythm, atrial fibrillation, and atrial flutter.
\vspace{0.3cm}  

\section{Methods}
\subsection{Preprocessing} 


Preprocessing of ECG signals is critical to ensure consistent input to uPVC-Net across diverse datasets, capturing relevant features while mitigating noise and variability. The pipeline (Figure \ref{fig:mainfigure}.b) involves the following steps, each designed to enhance robustness and generalization:

\textbf{Resampling}: ECG datasets in this study have varying native sampling rates (200–360 Hz). To ensure consistency, all signals are resampled to a common sampling rate of 200 Hz, the lowest native sampling rate among the datasets. This standardizes temporal resolution, minimizes interpolation artifacts, and reduces computational overhead during training. A sampling frequency of 200 Hz is sufficient to include the main frequency content of the ECG~\cite{cliffordadvanced}.

\textbf{Segmentation}: For each annotated heartbeat (referred to as a "reference beat") to be classified as PVC or non-PVC, an 8-second ECG segment is extracted and centered on it. This segment provides contextual information from surrounding beats and comprises 1600 samples (800 samples before and after the timestamp, at 200 Hz).

\textbf{Parametrization}: Each segment is transformed into a Mel-spectrogram using the \texttt{MelSpectrogram} function from Torchaudio (v2.0.2)~\cite{yang_torchaudio_2022}. This time-frequency representation captures both spectral and temporal features, crucial for identifying PVC morphologies that vary in shape and duration. We use 48 filterbanks spanning 0.5 to 40 Hz, which inherently band-pass-filters the signal by computing energy only within this range, eliminating the need for separate filtering. This frequency range retains key ECG components while attenuating low-frequency baseline wander and high-frequency noise. An FFT size of 256 with a hop size of 128 samples (50\% overlap between adjacent FFT windows) balances temporal resolution and computational efficiency, producing 13 time frames. The first and last frames are discarded to avoid edge windowing artifacts, resulting in an 11-frame representation. Note that the high filter density relative to the ECG’s bandwidth makes the Mel-spectrogram approximate a near-linear frequency scale, suitable for physiological signals.

\textbf{Standardization}: Spectrogram magnitudes often span a wide dynamic range due to variations in signal amplitude across recordings. A base-10 logarithmic transformation compresses this range, emphasizing relative energy differences and enhancing robustness to amplitude variations, as is standard in signal processing. Subsequently, channel-wise batch normalization is applied to each segment’s spectrogram, standardizing each of the 48 filterbanks independently across the 11 time frames, ensuring zero mean and unit variance for each frequency band. 

To ensure high-quality data for training and fair evaluation, specific data exclusions are applied. Unlabeled beats, ranging from 0.02\% in INCART to 23.33\% in Icentia11k, are excluded from analysis to maintain annotation reliability. Additionally, to align with the methodology of Li et al.~\cite{25-li2019ventricular} for fair comparison, the first and last three seconds of each MIT-BIH record are excluded.

\subsection{Training}
uPVC-Net employs a Bidirectional Gated Recurrent Unit (Bi-GRU) backbone composed of two recurrent layers, which processes the spectral frames sequence of the 8-second ECG segment. The network adopts a sequence-to-one strategy, where the hidden state corresponding to the central time step is used as the global representation of the input, as it receives equal contextual support from both past and future frames within the window.  This output is subsequently passed through a fully connected two-layer classifier to produce the final class prediction (Table \ref{table:upvcnet_architecture}).

\begin{table}[htbp]
\centering
\begin{tabular}{ccc}
\hline
\textbf{Shape} & \textbf{Layer} & \textbf{Descriptions} \\
\hline \\[-0.8em]
$1600 \times 1$ & Raw data & 8-second ECG window\\
$48 \times 11$ & Input & Preprocessed Mel-spectogram \\
$48 \times 128$ & Bi-GRU & \\
$128 \times 1$ & Central output &\\
$64 \times 1$ & Dense &  \\
$64 \times 1$ & ReLU &  \\
$1 \times 1$ & Dense &  \\
$1 \times 1$ & Sigmoid & \\
\hline
\end{tabular}
\vspace{0.5em}
\caption{uPVC-Net architecture.}
\label{table:upvcnet_architecture}
\end{table}

The Bi-GRU was chosen for its ability to capture temporal dependencies across frames, with bidirectionality contributing for distinguishing PVCs based on surrounding morphology. This choice was further justified by a theoretical comparison with alternative architectures: Conv2D, as used in Li et al. ~\cite{25-li2019ventricular}, introduce greater complexity; Conv1D is less suited for bidirectional temporal modeling; vanilla RNNs may face vanishing gradient issues; LSTMs are computationally heavier; and Transformers are unnecessarily complex and data-intensive for our short input sequences.

The Bi-GRU weights are initialized using uniforms distributions $U\bigl(-\sqrt{k}, \sqrt{k}\bigr), \quad \text{where } k = 1/hidden\_size = 1/128$. The linear classifier head is initialized using the Kaiming Uniform method\cite{he_imagenet_2015}. Any other bias term is initialized as zero.

uPVC-Net is trained using a multi-source multi-lead training strategy, illustrated in Figure \ref{fig:mainfigure}.b. That is, all leads from all available datasets but one are used for training while OOD-GP is reported for individual leads of the left-out dataset. Of note, beats classified as low quality are discarded from the training set to ensure that only high-quality data contribute to model training. However, to maintain comparability with other reports, these beats are not excluded from the target domain, and thus performance metrics are reported for the entire dataset.

To evaluate the impact of multi-source multi-channel training strategy, the proposed architecture is benchmarked on both channels of the MIT-BIH dataset. In each case, it is trained on an equal number of examples drawn either from a single source domain or from multiple sources (Figure \ref{fig:training_strategy}). 

During training, a weighted loss was implemented to account for the class imbalance present in the training set.

\subsection{Performance statistics}
Open datasets often exhibit highly imbalanced class distributions and may not accurately represent the intended clinical use case population. Consequently, defining a fixed probability threshold to provide a sensitivity (Se) - specificity (Sp) tradeoff or proportion-dependent performance statistics is not the most appropriate approach. The receiver operating characteristic curve (AUROC), on the other hand, is a more natural measure for evaluating and comparing a model's ability to discriminate between two classes. Therefore, the primary performance statistic we use in this research is the AUROC, which is calculated per channel. To estimate the confidence interval for AUROC, we apply bootstrapping, resampling the test set 100 times.

With the exception of Table IV, all performance metrics - including Se, Sp, F1 score, and Matthews Correlation Coefficient (MCC) - are computed using a default probability threshold of 0.5. However, prior work commonly uses undisclosed or varying thresholds\cite{24-llamedo2011heartbeat,36-teijeiro2018heartbeat,25-li2019ventricular,26-cai2024hybrid}, and typically does not provide AUROC, which is threshold-independent. This limits direct comparison at a fixed threshold.
To enable a fair comparison with Li et al.\cite{25-li2019ventricular}, whose threshold is not reported, we additionally evaluate our model at an adjusted threshold of 0.9, which aligns the sensitivity-specificity balance with their reported results. This adjustment is used only for that comparison (Table IV).



\subsection{Ablation study}
To assess the impact of preprocessing steps on uPVC-Net performance, an ablation study was conducted. For this purpose, uPVC-Net was successively retrained while omitting first the band-pass filter, then also omitting the Bi-GRU component. For each model variant, performance is reported as the median AUROC aggregated across all leads from each target dataset. Statistical significance in AUROC differences between model variants, was assessed using the DeLong test, with significance defined as $p < 0.05$.

\subsection{Error analysis}
Since the Non-PVC class represents multiple reference labels in each dataset, and in order to investigate the instances where uPVC-Net classified beats incorrectly, we show the distribution of each predicted class, based on the reference labels of each dataset. Odds is defined per label, as its probability to yield a positive prediction (PVC), divided by its probability to yield a negative prediction (Non-PVC). This is done to facilitate comparison between labels of varying prevalence across the dataset, and to highlight those which are more prone for misclassification. To further refine the analysis, a subset of 6 FN beats with the lowest predicted score, were manually reviewed by a medical doctor (first author HH) and categorized into incorrect reference labels or other factors.

\section{Results}
\label{sec:results}

\begin{figure*}[htbp]
    \centering
    \includegraphics[width=1\linewidth]{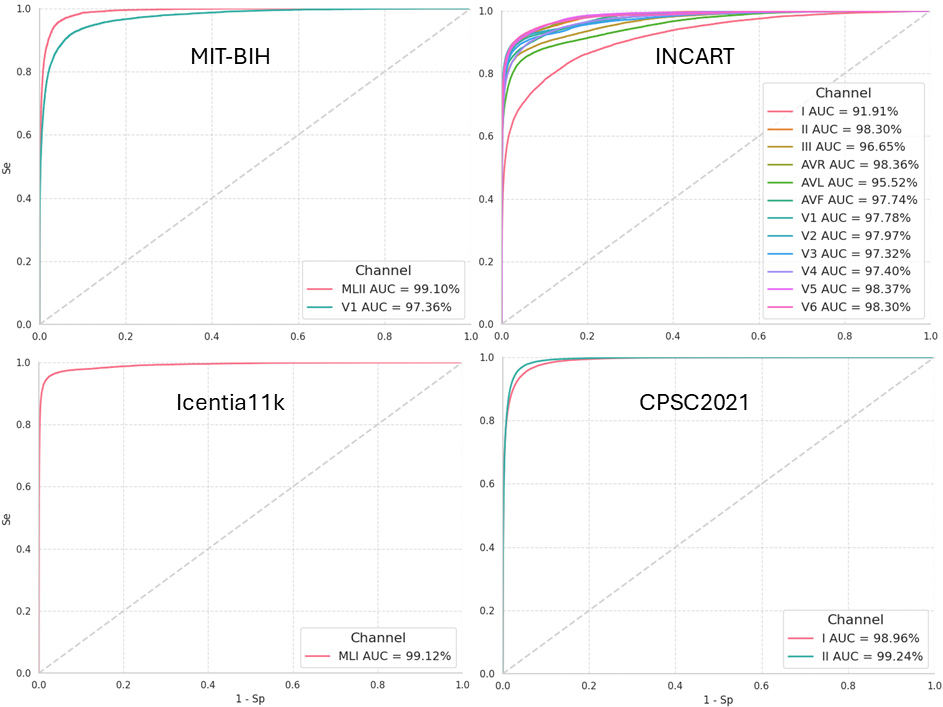}
    \caption{Receiver operating characteristic (ROC) curves for each channel in each dataset.}
    \label{fig:roc}
\end{figure*}

\begin{table}[]

\centering
\noindent\textbf{}
\vspace{0.5em}

\resizebox{\linewidth}{!}{%
    \begin{tabular}{@{}lccccc@{}}
    \toprule

\setlength{\tabcolsep}{0.5pt} 

 & \textbf{MIT-BIH} & \textbf{Icentia11k} & \textbf{INCART} & \textbf{CPSC2021} \\
\midrule
\small AUROC & \small 98.23 & \small 99.12 & \small 97.76 & \small 99.10 \\
(\%) & 98.11--98.37 & 98.98--99.21 & 97.66--97.87 & 99.07--99.12 \\
\midrule
\small Se & \small 91.64 & \small 89.55 & \small 87.92 & \small 87.35 \\
(\%) & 91.09--92.34 & 88.88--90.21 & 87.41--88.36 & 87.06--87.64 \\
\midrule
\small Sp & \small 95.64 & \small 99.39 & \small 97.23 & \small 98.42 \\
(\%) & 95.50--95.76 & 99.38--99.40 & 97.15--97.30 & 98.40--98.43 \\
\midrule
\small F1 & \small 73.10 & \small 66.33 & \small 82.89 & \small 68.07 \\
(\%) & 72.29--73.85 & 65.73--66.98 & 82.56--83.30 & 67.79--68.38 \\
\midrule
\small MCC & \small 0.724 & \small 0.684 & \small 0.796 & \small 0.689 \\
& 0.719--0.730 & 0.678--0.690 & 0.794--0.797 & 0.687--0.691

\\ \bottomrule
\end{tabular}%
}
\caption{uPVC-Net out-of-distribution generalization performance. Range indicates 95\% confidence interval.}
\label{tab:model_performance}
\end{table}

\subsection{uPVC-Net performance}
uPVC-Net achieved an AUROC of 97.76–99.12\% in OOD-GP over four different datasets. Other metrics are detailed in Table \ref{tab:model_performance} for a default probability threshold of 0.5. Figure \ref{fig:roc} shows the ROC curves for each channel in each dataset.

\begin{figure}[htbp]
    \centering
    \includegraphics[width=1\linewidth]{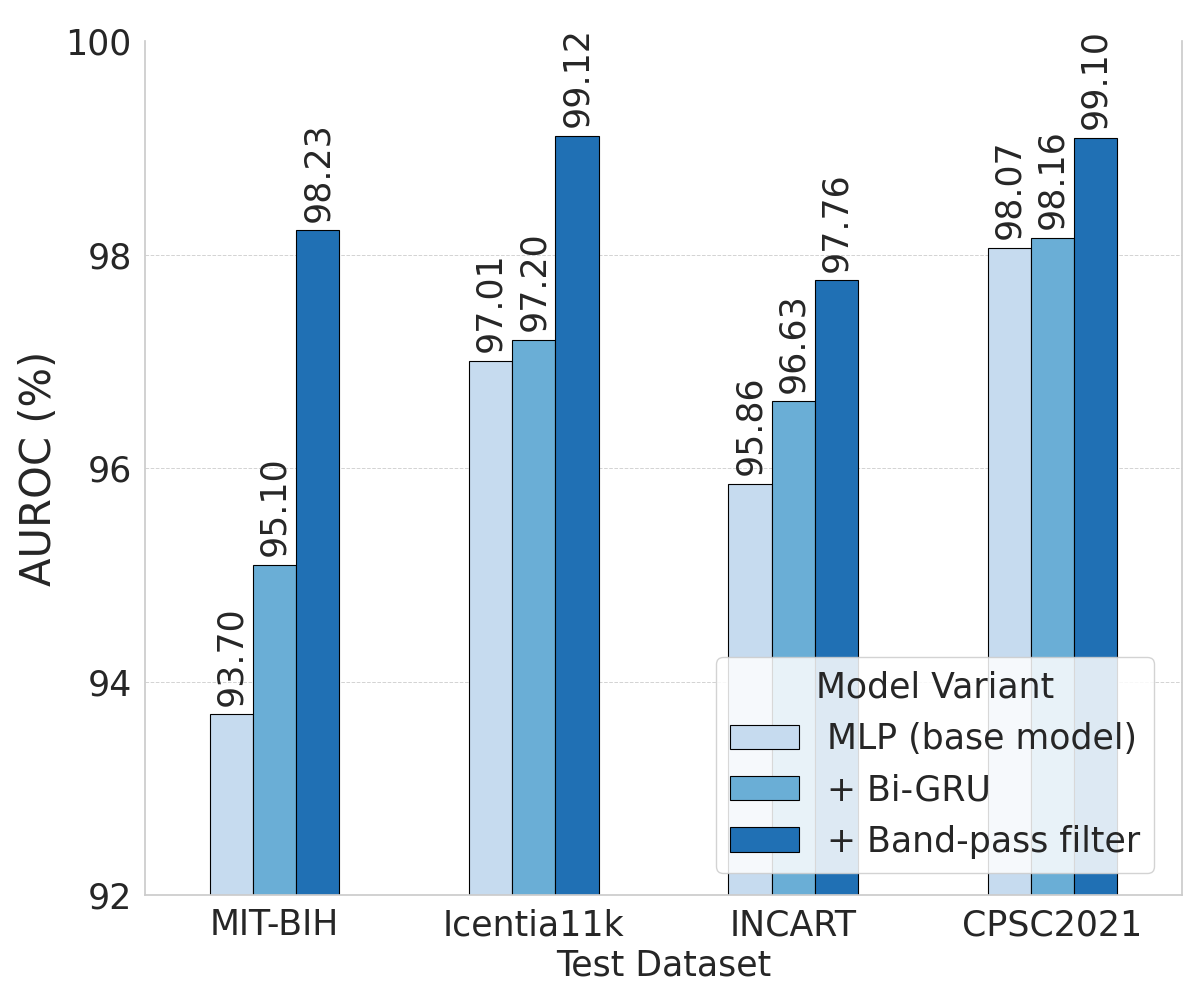}
    \caption{Ablation study. uPVC-Net was successively retrained using the same multi-source multi-lead training strategy, while first omitting the band-pass filter, then also omitting the Bi-GRU layer. OOD-GP is displayed for the left-out dataset.}
    \label{fig:ablation}
\end{figure}

The ablation study was conducted to assess the contribution of key preprocessing steps and model components (Figure \ref{fig:ablation}). The results indicate that introducing a Bi-GRU component significantly increases the AUROC by 0.09-1.40\% ($p-value < 0.001$). Applying a band-pass filter improves the AUROC by additional 0.94-3.13\%, an improvement which was very significant for all individual datasets ($p-value < 0.001$), except for the smallest dataset, MIT-BIH ($p-value = 0.053$). Finally, training on data drawn from multiple sources outperformed the single source training strategy in all data regimens (Figure \ref{fig:training_strategy}).

\subsection{Comparison to benchmark}
\label{subsec:comparison-to-benchmarks}

\begin{table}[tbp]
\centering
\noindent\textbf{}
\resizebox{\columnwidth}{!}{%
    \begin{tabular}{@{}lcc@{}}
    \toprule
\textbf{} & \textbf{uPVC-Net} & \textbf{Li et al.} \\ \midrule
Accuracy (\%) & 97.83 & 97.56\\
& 97.75 – 97.91 &\\
\midrule
Sensitivity (\%) & 85.23 & 82.55\\
&  84.48 – 86.00 &\\
\midrule
Specificity (\%) & 98.76 & 98.68\\
&  98.69 – 98.82 &\\
\midrule
PPV (\%) & 83.46 & 82.39\\
& 82.65 – 84.27  &\\
\midrule
NPV (\%) & 98.91 & -\\
&  98.84 – 98.97 &\\
\midrule
F1 (\%) & 84.33 & 82.47\\
&  83.78 – 84.90 &\\
\midrule
AUROC (\%) & 99.10 & -\\
&  99.03 – 99.17 &
\\ \bottomrule
\end{tabular}%
}

\caption{Comparison to Li et al.\cite{25-li2019ventricular} All performance measures relate to the first channel in MIT-BIH dataset. uPVC-Net applied with a classification threshold of 0.9 to allow clear comparison. Range indicates 95\% confidence interval. Negative predictive value (NPV) and AUROC were not reported by Li et al.}
\label{tab:qichen_benchmark}
\end{table}

uPVC-Net significantly outperforms the results reported by Li et al. \cite{25-li2019ventricular} in all performance measures, as shown in Table \ref{tab:qichen_benchmark}. In terms of computational complexity, uPVC-Net is significantly simpler than \cite{25-li2019ventricular}, both in signal representation and model architecture. While both approaches transform 1D ECG signals into 2D time-frequency representations, \cite{25-li2019ventricular} employs a multi-scale continuous wavelet transform (CWT) to generate its input images, whereas uPVC-Net relies on a single spectral-temporal representation via short-time Fourier transform (STFT). The use of multiple scales in the CWT increases parametrization complexity approximately linearly with the number of scales - in this case, involving 41 scales. Furthermore, the model architecture used in \cite{25-li2019ventricular}, based on Conv2d layers, contains over ten times more trainable parameters than our lightweight Bi-GRU based network. When considering full inference complexity - including both feature extraction and model processing - our system requires approximately 20× less FLOPS per ECG signal.


\begin{table}[tbp]
\centering

\begin{minipage}[t]{\linewidth}
\resizebox{\linewidth}{!}{%
\centering
{\textbf{MIT-BIH}\par\vspace{0.3em}}
\vspace{0.3em}
\begin{adjustbox}{max width=\linewidth}
\begin{tabular}{crrc}
\toprule

& \multicolumn{2}{c}{uPVC-Net} \\[0.3em]

Label & Non-PVC & PVC & Odds\\
\midrule
$S$ & 4 & 0 & 0.000\\
$R$ & 14,191 & 279 & 0.020\\
$j$ & 443 & 15 & 0.034\\
$N$ & 143,640 & 4,938 & 0.034\\
$L$ & 15,560 & 544 & 0.035\\
$J$ & 152 & 14 & 0.092\\
$e$ & 29 & 3 & 0.103\\
$A$ & 4,168 & 904 & 0.217\\
$F$ & 554 & 1,048 & 1.892\\
$a$ & 65 & 235 & 3.615\\
$E$ & 44 & 168 & 3.818\\
$V$ & 1,152 & 12,620 & 10.955\\

\bottomrule
\vspace{0.15em}
\end{tabular}
\end{adjustbox}
}
\end{minipage}
\hfill
\begin{minipage}[t]{\linewidth}
\resizebox{\linewidth}{!}{%
\centering

{\textbf{Icentia11k}\par\vspace{0.3em}}
\vspace{0.3em}
\begin{adjustbox}{max width=\linewidth}
\begin{tabular}{crrc}
& \multicolumn{2}{c}{uPVC-Net} \\[0.3em]
Label & Non-PVC & PVC & Odds\\
\midrule
$N$ & 1,298,470 & 6,868 & 0.005\\
$S$ & 10,946 & 1,166 & 0.107\\
$V$ & 1,043 & 8,939 & 8.570\\

\bottomrule
\vspace{0.15em}
\end{tabular}
\end{adjustbox}
}
\end{minipage}

\hfill
\begin{minipage}[t]{\linewidth}
\resizebox{\linewidth}{!}{%
\centering

{\textbf{INCART}\par\vspace{0.3em}}
\vspace{0.3em}
\begin{adjustbox}{max width=\linewidth}
\begin{tabular}{crrc}
& \multicolumn{2}{c}{uPVC-Net} \\[0.3em]
Label & Non-PVC & PVC & Odds\\
\midrule
$n$ & 379 & 5 & 0.013\\
$j$ & 1,087 & 17 & 0.016\\
$N$ & 1,761,482 & 43,390 & 0.025\\
$R$ & 35,851 & 2,225 & 0.062\\
$S$ & 151 & 41 & 0.272\\
$A$ & 15,267 & 8,061 & 0.528\\
$F$ & 1,194 & 1,434 & 1.201\\
$V$ & 35,298 & 204,834 & 5.803\\

\bottomrule
\vspace{0.15em}
\end{tabular}
\end{adjustbox}
}
\end{minipage}

\vspace{1em}

\begin{minipage}[t]{\linewidth}
\resizebox{\linewidth}{!}{%
\centering

{\textbf{CPSC2021}\par\vspace{0.3em}}
\vspace{0.3em}
\begin{adjustbox}{max width=\linewidth}
\begin{tabular}{crrc}
& \multicolumn{2}{c}{uPVC-Net} \\[0.3em]
Label & Non-PVC & PVC & Odds\\
\midrule
$a$ & 7,740 & 92 & 0.012\\
$N$ & 4,009,033 & 49,515 & 0.012\\
$A$ & 67,669 & 14,873 & 0.220\\
$F$ & 170 & 220 & 1.294\\
$E$ & 546 & 1,068 & 1.956\\
$V$ & 11,948 & 82,514 & 6.906\\

\bottomrule
\end{tabular}
\end{adjustbox}
}
\end{minipage}

\caption{uPVC-Net predictions per reference annotation in each dataset. Reference beat labels: $A$ – Atrial premature beat;
$E$ – Ventricular escape beat;
$F$ – Fusion of ventricular and normal beat;
$J$ – Nodal (junctional) premature beat;
$L$ – Left bundle branch block (LBBB) beat;
$N$ – Normal beat;
$R$ – Right bundle branch block (RBBB) beat;
$S$ – Supraventricular premature beat;
$V$ – Premature ventricular contraction (PVC);
$a$ – Aberrated atrial premature beat;
$e$ – Atrial escape beat;
$j$ – Nodal (junctional) escape beat;
$n$ - Supraventricular escape beat.}
\label{tab:beat_distributions}
\end{table}

\subsection{Error analysis}
\begin{figure*}[htbp]
    \centering
    \includegraphics[width=1\linewidth]{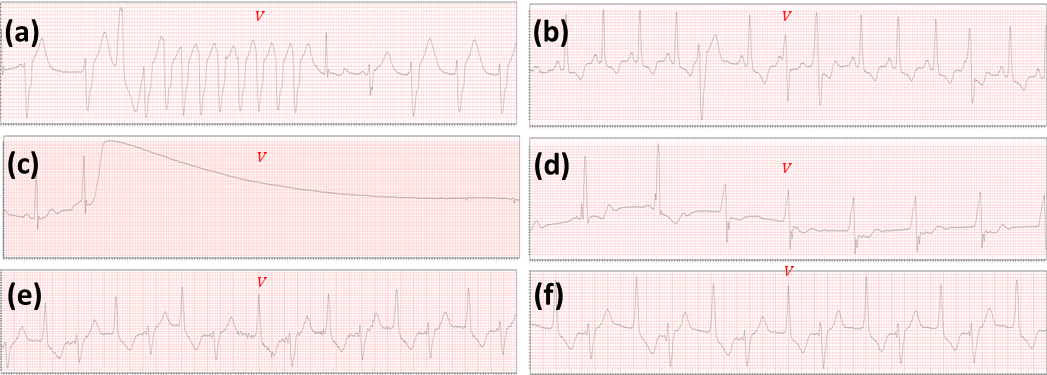}
    \caption{Six 8-second ECG segments centered around a FN beat (marked by a red $V$) with the lowest predicted PVC probability by uPVC-Net, among all test samples in the first channel of the MIT-BIH dataset. (a) Nonsustained ventricular tachycardia. (b) Fusion beat. (c) Artifact. (d) Ventricular escape beat. (e + f) Bigeminy with LBBB.}
    \label{fig:fn}
\end{figure*}

Among FP cases, the majority of beats - between 61\% and 86\% - had a normal beat reference ($N$). Fusion beats ($F$), aberrated premature atrial beats ($a$) and escape ventricular beats ($E$) were up to 1.9, 3.6 and 3.8 times more likely to be misclassified as PVC, respectively (Table. \ref{tab:beat_distributions}). Among the 6 FN cases with lowest PVC probability in the first channel of the MIT-BIH dataset, 3 were due to incorrect reference labels (fusion beat, artifact and ventricular escape beat); 2 cases were in a bigeminy sequence mixed with LBBB; and 1 was located in the middle of nonsustained ventricular tachycardia (Figure \ref{fig:fn}).


\begin{figure}[htbp]
    \centering
    \includegraphics[width=1\linewidth]{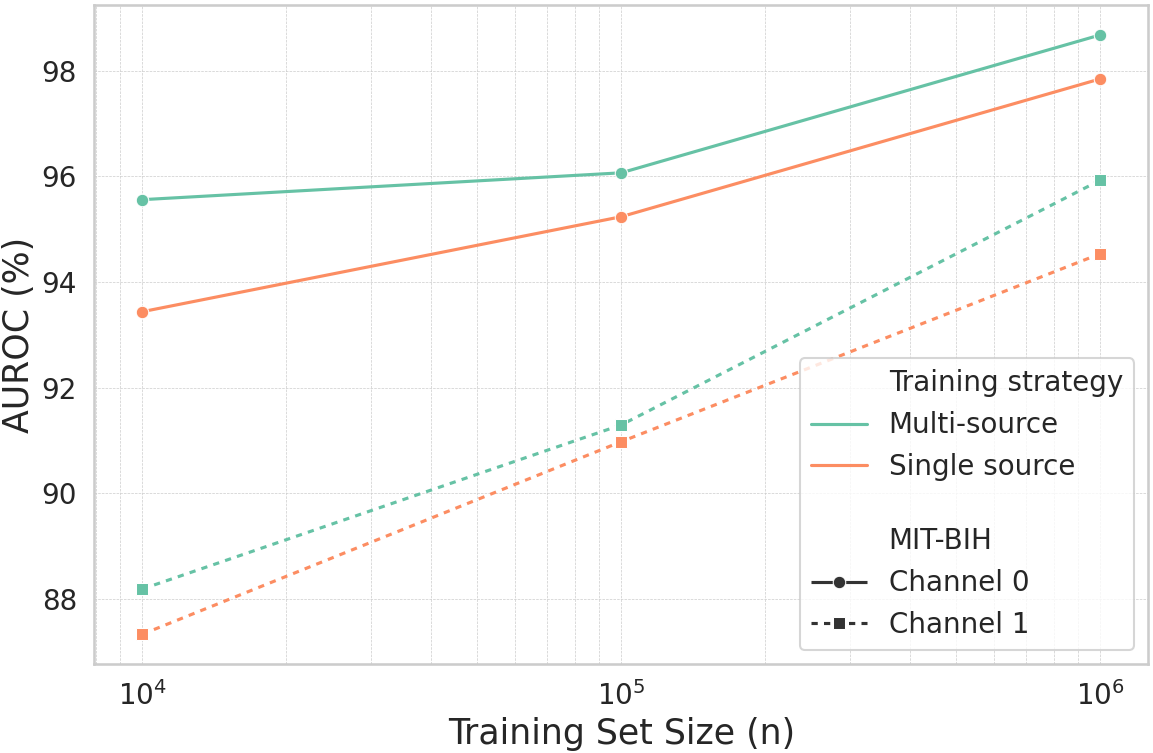}
    \caption{Training strategy. uPVC-Net was successively retrained using either $n$ training examples randomly drawn from a single source domain, or a total of $n$ training examples randomly drawn from three source domains. OOD-GP is then evaluated over both channels of MIT-BIH. For the single source training strategy the median AUROC over models trained on one dataset among Icentia11k, INCART or CPSC2021 is reported.}
    \label{fig:training_strategy}
\end{figure}

\section{Discussion}
\label{sec:discussion}

This study introduced uPVC-Net, a deep learning model designed for universal PVC beat detection from single-lead raw ECG recordings. By leveraging a multi-source domain, multi-lead training strategy, the model demonstrated strong generalization capabilities across diverse datasets, lead positions, and population samples, addressing a key limitation in deep learning-based PVC OOD-GP. Experimental evaluations conducted on four standard independent ECG datasets, comprising 8.3 million beats, validated the robustness of uPVC-Net. This research reports the best OOD-GP for the open datasets included in our experimental settings.

Several recent and significant studies have explored deep learning for PVC detection. Acharya et al. (2017) reported a higher performance on the MIT-BIH dataset than most previous and subsequent work\cite{34-acharya2017deep}. However, as noted by Li et al. (2019)\cite{25-li2019ventricular}, their study suffered from information leakage, as the beats were divided by class rather than by patient. Li et al. reported an overall F1 score of 84.94\% on their local MIT-BIH test set and 84.96\% on the American Heart Association (AHA) dataset, which they considered a target domain. The AHA dataset is not publicly available. Li et al. then proceeded to retrain their model on the AHA dataset and re-evaluate it on the MIT-BIH dataset, as an external test set, albeit their system was originally calibrated to the MIT-BIH dataset. More recently, Cai et al. (2024) employed a graph neural network approach for PVC identification, achieving an F1 score of 93.06\% on INCART and 97.02\% on MIT-BIH\cite{26-cai2024hybrid}. However, in their experimental setup, the authors included only PVC and normal beats, which inherently exclude all other ectopic beats, e.g. premature atrial contractions and fusion beats. This limitation does not reflect a real-world clinical scenario. In addition, some previous studies involved automated\cite{17-cai2022robust} or semi-automated\cite{oster_semisupervised_2015} initialization period towards the elaboration of template beats. Our approach requires no initialization, is fully automated, proven to work with both classical and non-classical lead systems.

The approach presented in this research leverages both a multi-source, multi-channel training strategy, which significantly improves OOD-GP by up to 2.1\% (Figure \ref{fig:training_strategy}), and a well-designed preprocessing pipeline (Figure \ref{fig:mainfigure}.b.), which further enhances OOD-GP by 2.4\% on average. This comprehensive approach enabled uPVC-Net to achieve robust OOD-GP, with AUROC ranging from 97.76\% to 99.12\%. Notably, uPVC-Net demonstrated strong performance on single-lead ECG recordings from both traditional Holter devices (MIT-BIH, INCART, CPSC) and a modern single-lead ECG wearable (Icentia11k). This suggests that uPVC-Net is well-suited for deployment on modern ECG patches and wearables, which are often characterized by lower signal quality compared to conventional Holter monitors.

Beyond the specific focus on PVC detection, our research highlights the feasibility of developing universal deep learning models for physiological time series analysis. These models are designed to operate across data recorded by various sensors from different manufacturers and electrode placements. This perspective aligns with our vision for the field\cite{behar_generalization_2025} but contrasts with the prevailing belief that models must be adapted to specific target domains to achieve high performance\cite{futoma_generalisability_2020,youssef_externalvalidation_2023}.

Further improvements to the model could be achieved by performing a thorough hyperparameter search over window size used for the initial ECG strip segmentation step or the batch size used. Replacing the GRU with a simpler RNN architecture may further improve since the current context length is relatively short, and the added complexity of GRUs may be unnecessary. This could further reduce the size of the model and computational cost. Additionally, temporal and cross-frequency attention mechanisms over the filterbank sequences may enhance pattern recognition capabilities. Another existing perspective for improvement is the usage of ECG foundation models \cite{coppola_hubert-ecg_2024,li_electrocardiogram_2024} which have shown interesting results in boosting various ECG tasks. Error analysis reveals certain types of beats - namely fusion beats ($F$) and ventricular escape beats ($E$) - are prone for misclassification, suggesting augmenting these specific beats may further improve performance. Finally, inline with both the current work and the work of Zvuloni et al. \cite{eran_multisource_2025}, increasing the number of source domains may be key to enhancing OOD-GP.

Our research has several important limitations. First, it includes a limited number of independent datasets (n=4), which inherently restricts the geographical diversity and ethnic representation of the study population. Thus, future studies should aim to include populations from Africa, South and Southeast Asia, Latin America, the Middle East, and Oceania. Second, we relied on reference beat locations to segment and classify each beat. Some of these datasets have been curated, meaning that any beat misdetections by the QRS detector may have been manually corrected. As a result-similar to other studies-our findings may be slightly inflated. However, given the high accuracy\cite{llamedo_qrs_nodate} of modern beat detection algorithms, we expect this effect to be minimal. Moreover, alternative deep learning architectures, particularly modern transformer-based models, could be evaluated and benchmarked. We also note that beat annotations across the different datasets were performed by different annotators, which may introduce inconsistencies due to varying criteria or levels of expertise. Additionally, the robustness of uPVC-Net under varying levels of noise remains to be systematically evaluated. Finally, beyond the error analysis conducted in our study, it would be valuable to further investigate whether misclassifications of abnormal beats such as F or E are isolated or sequential, and whether they occur during arrhythmic events.

\section{Conclusion}
\label{sec:conclusion}
This study introduced uPVC-Net, a novel deep learning model for the automatic detection of PVC from ECG signals acquired across diverse datasets, lead configurations, and recording devices. The proposed Bi-GRU architecture has demonstrated robust out-of-distribution generalization performance, achieving an AUROC of 97.76\%-99.12\% and surpassing a state-of-the-art model.
The universality of uPVC-Net stems from its multi-source domain training strategy, which enables effective adaptation to heterogeneous ECG sources, regardless of number of leads or lead configuration. This characteristic supports its application in various clinical and research contexts, including short ECG recordings, extended Holter monitoring, and near-real-time analysis in wearable or bedside devices.
These findings highlight the potential of uPVC-Net as a versatile and scalable tool for automated PVC screening and continuous cardiac monitoring. Future work will evaluate its integration into real-world clinical workflows and investigate its capability to predict broader cardiac risk profiles associated with PVC occurrence.

\section{References}
\bibliographystyle{ieeetr}
\bibliography{zotero}

@article{dukes_pvc_HF_death_2015,
author={Jonathan W. Dukes and others},
year={2016},
month={Jul},
title={Ventricular Ectopy as a Predictor of Heart Failure and Death},
journal={J Am Coll Cardiol},
volume={66},
number={2},
pages={101–109},
isbn={0735-1097},
doi={10.1016/j.jacc.2015.04.062}
}

@article{agarwal_pvc_stroke_2011,
author={Sunil K. Agarwal and others},
year={2010},
month={Apr},
title={Premature ventricular complexes and risk of incident stroke: The
Atherosclerosis Risk In Communities ({ARIC}) study},
journal={Stroke},
volume={41},
number={4},
pages={588–593},
isbn={0039-2499},
doi={10.1161/strokeaha.109.567800}
}

@article{agarwal_pvc_stroke_2016,
author={Sunil K. Agarwal and others},
year={2015},
month={May},
title={Premature Ventricular Complexes on Screening Electrocardiogram and Risk of Ischemic Stroke},
journal={Stroke},
volume={46},
number={5},
pages={1365–1367},
isbn={0039-2499},
doi={10.1161/strokeaha.114.008447}
}

@article{yu_multiform_pvc_2014,
author={Chin-Yu Lin and others},
year={2014},
month={Nov},
title={Long-term outcome of multiform premature ventricular complexes in structurally normal heart},
journal={Int J Cardiol},
volume={180},
pages={80-85},
isbn={0167-5273},
doi={10.1016/j.ijcard.2014.11.110}
}

@article{baman2010relationship,
	author = 	 {Timir S. Baman and others},
	year = 	 {2010},
	month = 	 {Jul},
	title = 	 {Relationship between burden of premature ventricular complexes and left ventricular function},
	journal = 	 {Heart Rhythm},
	volume = 	 {7},
	number = 	 {7},
	pages = 	 {865},
	isbn = 	 {1547-5271},
	doi={10.1016/j.hrthm.2010.03.036}
}

@article{penela2017clinical,
	author = 	 {Diego Penela and others},
	year = 	 {2017},
	month = 	 {Dec},
	title = 	 {Clinical recognition of pure premature ventricular complex-induced cardiomyopathy at presentation},
	journal = 	 {Heart Rhythm},
	volume = 	 {14},
	number = 	 {12},
	pages = 	 {1864},
	isbn = 	 {1547-5271},
	doi={10.1016/j.hrthm.2017.07.025}
}

@article{latchamsetty2015multicenter,
	author = 	 {Rakesh Latchamsetty and others},
	title = 	 {Multicenter Outcomes for Catheter Ablation of Idiopathic Premature Ventricular Complexes},
journal={JACC Clin Electrophysiol},
year={2015},
month={Jun},
volume = 	 {1},
	number = 	 {3},
	pages = 	 {116-123}
}

@article{bogun2007radiofrequency,
	author = 	 {Frank Bogun and others},
	year = 	 {2007},
	month = 	 {Jul},
	title = 	 {Radiofrequency ablation of frequent, idiopathic premature ventricular complexes: Comparison with a control group without intervention},
	journal = 	 {Heart Rhythm},
	volume = 	 {4},
	number = 	 {7},
	pages = 	 {863},
	isbn = 	 {1547-5271},
	doi={10.1016/j.hrthm.2007.03.003}
}

@article{12-cai2019rule-based,
author={Zhipeng Cai and others},
year={2019},
month={Oct},
title={Rule-based rough-refined two-step-procedure for real-time premature beat detection in single-lead {ECG}},
journal={Physiol Meas},
volume={41},
number={5},
pages={054004},
doi={10.1088/1361-6579/ab87b4}
}

@article{9-hajeb-mohammadalipour2018automated,
author={Shirin Hajeb-Mohammadalipour and others},
year={2018},
month={Jun},
title={Automated Method for Discrimination of Arrhythmias Using Time, Frequency, and Nonlinear Features of Electrocardiogram Signals},
journal={Sensors},
volume={18},
number={7},
pages={2090},
doi={10.3390/s18072090}
}

@article{8-oliveira2019geometrical,
author={Bruno Rodrigues De Oliveira and others},
year={2019},
month={Feb},
title={Geometrical features for premature ventricular contraction recognition with analytic hierarchy process based machine learning algorithms selection},
journal={Comput Methods Programs Biomed},
volume={169},
pages={59},
isbn={0169-2607},
doi={10.1016/j.cmpb.2018.12.028}
}

@article{7-de2022classification,
author={Fabiola De Marco and others},
year={2022},
month={Aug},
title={Classification of QRS complexes to detect Premature Ventricular Contraction using machine learning techniques},
journal={PLoS One},
volume={17},
number={8},
pages={e0268555},
doi={10.1371/journal.pone.0268555}
}

@article{10-sarshar2022premature,
author={Nazanin Tataei Sarshar and Mohammad Mirzaei},
year={2022},
month={Mar},
title={Premature Ventricular Contraction Recognition Based on a Deep Learning Approach},
journal={J Healthc Eng},
volume={2022},
pages={1},
isbn={2040-2295},
doi={10.1155/2022/1450723}
}

@article{17-cai2022robust,
author={Zhipeng Cai and others},
year={2022},
month={Mar},
title={Robust {PVC} Identification by Fusing Expert System and Deep Learning},
journal={Biosensors},
volume={12},
number={4},
pages={185},
doi={10.3390/bios12040185}
}

@article{18-ullah2022automatic,
author={Hadaate Ullah and others},
year={2022},
month={Dec},
title={An Automatic Premature Ventricular Contraction Recognition System Based on Imbalanced Dataset and Pre-Trained Residual Network Using Transfer Learning on {ECG} Signal},
journal={Diagnostics},
volume={13},
number={1},
pages={87},
doi={10.3390/diagnostics13010087}
}

@article{19-kraft2023end-to-end,
author={Dimitri Kraft and others},
year={2023},
month={Oct},
title={End-to-End Premature Ventricular Contraction Detection Using Deep Neural Networks},
journal={Sensors},
volume={23},
number={20},
pages={8573},
doi={10.3390/s23208573}
}

@article{11-guo2024pvcsnet,
author={Biren Guo and others},
year={2024},
month={Sep},
title={{PVCsNet}: A Specialized Artificial Intelligence-Based Model to Classify Premature Ventricular Contractions from {ECG} Images},
journal={IEEE J Biomed Health Inform},
pages={1},
isbn={2168-2194},
doi={10.1109/jbhi.2024.3471510}
}

@article{24-llamedo2011heartbeat,
author={Mariano Llamedo and Juan Pablo Martínez},
year={2011},
month={Mar},
title={Heartbeat Classification Using Feature Selection Driven by Database Generalization Criteria},
journal={IEEE Trans Biomed Eng},
volume={58},
number={3},
pages={616},
isbn={0018-9294},
doi={10.1109/tbme.2010.2068048}
}

@article{25-li2019ventricular,
author={Qichen Li and others},
year={2019},
month={Jun},
title={Ventricular ectopic beat detection using a wavelet transform and a convolutional neural network},
journal={Physiol Meas},
volume={40},
number={5},
pages={055002},
doi={10.1088/1361-6579/ab17f0}
}

@article{26-cai2024hybrid,
author={Zhipeng Cai and others},
year={2024},
month={Jun},
title={Hybrid Amplitude Ordinal Partition Networks for {ECG} Morphology Discrimination: An Application to {PVC} Recognition},
journal={IEEE Trans Instrum Meas},
volume={73},
pages={1},
isbn={0018-9456},
doi={10.1109/tim.2024.3400307}
}

@article{phan_seqsleepnet_2019,
	author = 	 {Huy Phan and others},
	year = 	 {2019},
	month = 	 {Mar},
	title = 	 {{SeqSleepNet}: End-to-End Hierarchical Recurrent Neural Network for Sequence-to-Sequence Automatic Sleep Staging},
	journal = 	 {IEEE Trans Neural Syst Rehabil Eng},
	volume = 	 {27},
	number = 	 {3},
	pages = 	 {400},
	isbn = 	 {1534-4320},
	doi={10.1109/tnsre.2019.2896659}
}

@article{levy_deep_2023,
	author = 	 {Jeremy Levy and others},
	year = 	 {2023},
	month = 	 {Aug},
	title = 	 {Deep learning for obstructive sleep apnea diagnosis based on single channel oximetry},
	journal = 	 {Nat Commun},
	volume = 	 {14},
	number = 	 {1},
pages={4881},
	doi={10.1038/s41467-023-40604-3}
}

@article{ribeiro_automatic_2020,
	author = 	 {Antônio H. Ribeiro and others},
	year = 	 {2020},
	month = 	 {Apr},
	title = 	 {Automatic diagnosis of the {12}-lead {ECG} using a deep neural network},
	journal = 	 {Nat Commun},
	volume = 	 {11},
	number = 	 {1},
pages={1760},
	doi={10.1038/s41467-020-15432-4}
}

@article{kotzen_sleepppg-net_2023,
	author = 	 {Kevin Kotzen and others},
	year = 	 {2023},
	month = 	 {Feb},
	title = 	 {{SleepPPG-Net}: A Deep Learning Algorithm for Robust Sleep Staging From Continuous Photoplethysmography},
	journal = 	 {IEEE J Biomed Health Inform},
	volume = 	 {27},
	number = 	 {2},
	pages = 	 {924},
	isbn = 	 {2168-2194},
	doi={10.1109/jbhi.2022.3225363}
}

@article{ballas_towards_2024,
	author = 	 {Aristotelis Ballas and Christos Diou},
	year = 	 {2024},
	month = 	 {Feb},
	title = 	 {Towards Domain Generalization for {ECG} and {EEG} Classification: Algorithms and Benchmarks},
	journal = 	 {IEEE Trans Emerg Top Comput Intell},
	volume = 	 {8},
	number = 	 {1},
	pages = 	 {44},
	doi={10.1109/tetci.2023.3306253}
}

@article{behar_generalization_2025,
author={Eran Zvuloni and others},
title={Generalization in medical {AI}: a perspective on developing scalable models},
year={2025},
month={Apr},
journal={arXiv:2311.05418v2}
}

@article{attia_sleepppg_2024,
author={Shirel Attia and others},
year={2024},
month={Apr},
title={{SleepPPG-Net2}: Deep learning generalization for sleep staging from photoplethysmography},
journal={arXiv:2404.06869}
}

@article{men_deep_2025,
	title = {Deep learning generalization for diabetic retinopathy staging from fundus images},
	author = {Men, Yevgeniy and others},
	year = {2025},
month={Jan},
journal={Physiol Meas},
volume={46},
number={1},
}

@article{abramovich_gonet_2025,
	title = {{GONet}: A Generalizable Deep Learning Model for Glaucoma Detection},
	author = {Abramovich, Or and others},
	year = {2025},
	doi = {10.1109/tbme.2025.3576688},
	note = {Accepted},
journal={IEEE Trans Biomed Eng},
}

@article{eran_multisource_2025,
	author={Eran Zvuloni and others},
	year={2025},
	title={A multi-source domain training framework for deep generalization performance in physiological time series analysis},
	journal={Under review in NPJ Digital Medicine}
}

@article{ben-moshe_rawecgnet_2024,
author={Noam Ben-Moshe and others},
year={2024},
month={Sep},
title={{RawECGNet}: Deep Learning Generalization for Atrial Fibrillation Detection From the Raw {ECG}},
journal={IEEE J Biomed Health Inform},
volume={28},
number={9},
pages={5180},
isbn={2168-2194},
doi={10.1109/jbhi.2024.3404877}
}

@article{physiobank,
author={Ary L. Goldberger and others},
year={2000},
month={Jun},
title={PhysioBank, PhysioToolkit, and PhysioNet Components of a New Research Resource for Complex Physiologic Signals},
journal={Circulation},
volume={101},
issue={23},
pages={e215–e220}
}

@article{mit_bih,
author={George B. Moody and Roger G. Mark},
title={The Impact of the {MIT-BIH} Arrhythmia Database},
journal={IEEE Eng Med Biol},
year={2001},
month={Jun},
volume={20},
number={3},
pages={45-50}
}

@article{icentia11k_ref2,
author={Shawn Tan and others},
title={Icentia11K: An Unsupervised Representation Learning Dataset for Arrhythmia Subtype Discovery},
note={arXiv:1910.09570, Oct 2019}
}

@article{icentia11k_ref1,
author={Tan, S. and others},
title={Icentia11k Single Lead Continuous Raw Electrocardiogram Dataset (version 1.0)},
year={2022},
month={Apr},
journal={PhysioNet},
note={https://doi.org/10.13026/kk0v-r952}
}

@article{CPSC2021,
author={Wang, X. and others},
title={Paroxysmal Atrial Fibrillation Events Detection from Dynamic {ECG} Recordings: The 4th {China} Physiological Signal Challenge 2021 (version 1.0.0)},
year={2021},
month={Jun},
journal={PhysioNet},
note={https://doi.org/10.13026/ksya-qw89}
}

@article{23-ec571998testing,
author={ANSI-AAMI EC57},
year={1999},
month={Jan},
title={Testing and reporting performance results of cardiac rhythm and {ST} segment measurement algorithms},
journal={Association for the Advancement of Medical Instrumentation, Arlington, VA}
}

@book{cliffordadvanced,
	author = 	 {Gari D. Clifford and others},
	title = 	 {Advanced Methods and Tools for {ECG} Data Analysis},
note={Boston, MA, USA: Artech House, 2006, ch. 7, p. 201.}
}

@article{yang_torchaudio_2022,
title={Torchaudio: Building Blocks for Audio and Speech Processing},
doi={10.1109/icassp43922.2022.9747236},
journal={IEEE},
author={Yang, Yao-Yuan and others},
month={May},
year={2022},
pages={6982},
}

@article{he_imagenet_2015,
author={Kaiming He and others},
title={Delving Deep into Rectifiers: Surpassing Human-Level Performance on {ImageNet} Classification},
year={2015},
month={Feb},
journal={IEEE Comput Soc},
pages={1026-1034}
}

@article{36-teijeiro2018heartbeat,
author={Tomas Teijeiro and others},
year={2018},
month={Mar},
title={Heartbeat Classification Using Abstract Features From the Abductive Interpretation of the {ECG}},
journal={IEEE J Biomed Health Inform},
volume={22},
number={2},
pages={409},
isbn={2168-2194},
doi={10.1109/jbhi.2016.2631247}
}

@article{34-acharya2017deep,
author={U. Rajendra Acharya and others},
year={2017},
month={Oct},
title={A deep convolutional neural network model to classify heartbeats},
journal={Comput Biol Med},
volume={89},
pages={389},
isbn={0010-4825},
doi={10.1016/j.compbiomed.2017.08.022}
}

@article{oster_semisupervised_2015,
	title = {Semisupervised {ECG} Ventricular Beat Classification With Novelty Detection Based on Switching {Kalman} Filters},
	volume = {62},
	doi = {10.1109/tbme.2015.2402236},
	number = {9},
	journal = {IEEE Trans Biomed Eng},
	author = {Oster, Julien and others},
	year = {2015},
month={Sep},
	note = {ISBN: 0018-9294},
	pages = {2125},
}

@article{futoma_generalisability_2020,
author={Joseph Futoma and others},
title={The myth of generalisability in clinical research and machine learning in health care},
year={2020},
month={Sep},
journal={Lancet Digit Health},
volume={2},
page={e489–e492}
}

@article{youssef_externalvalidation_2023,
author={Alexey Youssef and others},
year={2023},
month={Oct},
title={External validation of {AI} models in health should be replaced with recurring local validation},
journal={Nat Med},
volume={29},
number={11},
pages={2686},
isbn={1078-8956},
doi={10.1038/s41591-023-02540-z}
}

@article{coppola_hubert-ecg_2024,
author={Edoardo Coppola and others},
year={2024},
month={Nov},
title={{HuBERT-ECG}: a self-supervised foundation model for broad and scalable cardiac applications},
journal={medRxiv},
doi={10.1101/2024.11.14.24317328}
}

@article{li_electrocardiogram_2024,
author={Jun Li and others},
year={2025},
month={Apr},
title={An Electrocardiogram Foundation Model Built on over 10 Million Recordings with External Evaluation across Multiple Domains},
journal={arXiv:2410.04133v3}
}

@article{llamedo_qrs_nodate,
	title = {{QRS} Detectors Performance Comparison in Public Databases},
	author = {Llamedo, Mariano and others},
journal={Computing in Cardiology 2014},
year={2014},
month={Sep},
pages={357-360}
}

\end{document}